# Analysing Sensitivity Data from Probabilistic Networks

**Linda C. van der Gaag** and **Silja Renooij**
Institute of Information and Computing Sciences, Utrecht University
P.O. Box 80.089, 3508 TB Utrecht, The Netherlands
{linda,silja}@cs.uu.nl

## Abstract

With the advance of efficient algorithms for *sensitivity analysis* of probabilistic networks, studying the sensitivities revealed by real-life networks is becoming feasible. As the amount of data yielded by an analysis of even a moderately-sized network is already overwhelming, effective methods for extracting relevant information from these data are called for. One such method is to study the *derivatives* of the sensitivity functions yielded, to identify the parameters that upon variation are expected to have a large effect on a probability of interest. We further propose to build upon the concept of *admissible deviation*, which captures the extent to which a parameter can be varied without inducing a change in the most likely outcome. We illustrate these concepts by means of a sensitivity analysis of a real-life probabilistic network in the field of oncology.

## 1 Introduction

The numerical parameters for a probabilistic network are generally estimated from statistical data or assessed by human experts in the domain of application. As a result of incompleteness of data and partial knowledge of the domain, the assessments obtained inevitably are inaccurate. Since the outcome of a probabilistic network is built upon these assessments, it may be sensitive to the inaccuracies involved and, as a result, may even be unreliable.

The reliability of the outcome of a probabilistic network can be evaluated by subjecting the network to a *sensitivity analysis*. In general, sensitivity analysis of a mathematical model amounts to investigating the effects of inaccuracies in the model's parameters by systematically varying the values of these parameters [1]. For a probabilistic network, sensitivity analysis amounts to varying the assessments for one or more of its numerical parameters and investigating the effects on, for example, a probability of interest [2].

Straightforward sensitivity analysis of a probabilistic network is highly time-consuming. The probability of interest has to be computed from the network for a number of deviations from the original assessment for every single parameter. Even for a rather small network, this easily requires thousands of propagations. Recently, however, efficient algorithms for sensitivity analysis have become available, rendering analysis of real-life probabilistic networks feasible [2, 3]. These algorithms build upon the observation that the sensitivity of a probability of interest to parameter variation complies with a simple mathematical function; this *sensitivity function* basically expresses the probability of interest in terms of the parameter under study. Computing the constants for these sensitivity functions requires just a limited number of propagations.

Sensitivity analysis of a real-life probabilistic network tends to result in a huge amount of data. For example, for a real-life network in the field of oncology, comprising some 1000 parameters, we found that a single analysis yielded close to 5000 sensitivity functions with two or three constants each. Because the sensitivities exhibited by a probabilistic network typically change with evidence, such an analysis has to be performed a number of times. For our network, for example, we conducted over 150 analyses involving data from real patients. For extracting relevant information from the data thus generated, effective methods are called for that allow for (automatically) identifying parameters that are quite influential upon variation. These parameters should be selected for further investigation as the inaccuracies in their assessments are likely to affect the network's outcome.

In her work on sensitivity analysis of probabilistic networks, K. Blackmond Laskey introduced the concept of *sensitivity value* [4]. The concept builds upon the (partial) *derivative* of the sensitivity function that expresses the network's probability of interest in terms of a parameter under study: the sensitivity value is the absolute value of this derivative at the original assessment for the parameter. As currently available algorithms for sensitivity analysis of probabilistic networks yield the probability of in-terest explicitly as a function of the parameter under study, the derivative and its associated sensitivity value are readily determined. In this paper, we study the derivatives of sensitivity functions and show how sensitivity values can be used for selecting parameters for further investigation.

In many real-life applications of probabilistic networks, the outcome of interest is not a probability, but rather the most likely value of a variable of interest. In a medical application, for example, the outcome of interest may be the most likely diagnosis given a patient's symptoms and signs. For this type of outcome, the derivative of a sensitivity function and its associated sensitivity value are no longer appropriate for establishing the effect of parameter variation: a parameter with a large sensitivity value may upon variation not induce a change in the most likely outcome, while a parameter with a small sensitivity value may induce such a change for just a small deviation from its original assessment. To describe the sensitivities of this type of outcome, we introduce the concept of *admissible deviation*. This concept captures the extent to which a parameter can be varied without inducing a change in the most likely value for the variable of interest. We show how the concept of admissible deviation can be used for selecting parameters.

The various concepts introduced in this paper will be illustrated by means of a sensitivity analysis of a moderately-sized real-life probabilistic network in the field of oncology. We would like to note that this network exhibits considerable sensitivity to parameter variation. This observation contradicts earlier suggestions that probabilistic networks are highly insensitive to inaccuracies in the assessments for their parameters [5, 6]. Our results now seem to indicate that the sensitivities exhibited by probabilistic networks may vary from application to application.

The paper is organised as follows. In Section 2, we provide some preliminaries on sensitivity analysis of probabilistic networks. In Section 3, we briefly discuss the oesophagus network with which we will illustrate the concepts introduced in the subsequent sections. In Section 4, we study the derivative of a sensitivity function and its associated sensitivity value. In Section 5, we introduce the concept of admissible deviation and discuss its use for extracting relevant information from sensitivity data. The paper ends with our conclusions and directions for further research in Section 6.

## 2 Preliminaries

Sensitivity analysis of a probabilistic network amounts to establishing, for each of the network's numerical parameters, the *sensitivity function* that expresses the probability of interest in terms of the parameter under study. In the sequel, we denote the probability of interest by $\Pr(A = a \mid e)$, or $\Pr(a \mid e)$ for short, where $a$ is a specific value of the variable $A$ of interest and $e$ denotes the available evidence. The network's parameters are denoted by $x = p(b_i \mid \pi)$, where $b_i$ is a value of a variable $B$ and $\pi$ is a combination of values for the parents of $B$. We write $f_{\Pr(a\mid e)}(x)$ to denote the sensitivity function that expresses the probability $\Pr(a \mid e)$ in terms of the parameter $x$; for ease of exposition, we will often omit or abbreviate the subscript for the function symbol $f$, as long as ambiguity cannot occur.

Upon varying a single parameter $x = p(b_i \mid \pi)$ in a probabilistic network, the other parameters $p(b_j \mid \pi)$, $j \neq i$, specified for the variable $B$ need to be *co-varied*. Each parameter $p(b_j \mid \pi)$ can thus be seen as a function $g_j(x)$ of the parameter $x$ under study. In the sequel, we assume that the parameters $p(b_j \mid \pi)$ are co-varied in such a way that their mutual proportional relationship is kept constant, that is, a parameter $p(b_j \mid \pi)$ is co-varied according to

$$g_j(x) = \begin{cases} x & \text{if } j = i \\ p(b_j \mid \pi) \cdot \dfrac{1-x}{1-p(b_i \mid \pi)} & \text{otherwise} \end{cases}$$

for $p(b_i \mid \pi) < 1$.

Under the assumption of co-variation described above, a sensitivity function $f(x)$ is a quotient of two functions that are linear in the parameter $x$ under study [2, 7]; more formally, the function takes the form

$$f(x) = \frac{a \cdot x + b}{c \cdot x + d}$$

where the constants $a$, $b$, $c$, and $d$ are built from the assessments for the numerical parameters that are not being varied. From this property we have that a sensitivity function is characterised by at most three constants. These constants can be feasibly determined from the network, for example by computing the probability of interest for a small number of values for the parameter under study and solving the resulting system of equations [2]. An even more efficient algorithm that is closely related to junction-tree propagation, is available from [3].

## 3 The oesophagus network

The *oesophagus network* that will be used to illustrate concepts throughout this paper, is a real-life probabilistic network for the staging of oesophageal cancer. The network was constructed and refined with the help of two experts in gastrointestinal oncology from the Netherlands Cancer Institute, Antoni van Leeuwenhoekhuis, and is destined for use in clinical practice [8].

As a consequence of a lesion of the oesophageal wall, a tumour may develop in a patient's oesophagus. The characteristics of the tumour, such as its location in the oesophagus and its macroscopic shape, influence the tumour's prospective growth. The tumour typically invades the oesophageal wall and upon further growth may affect neigh-bouring organs. In time, the tumour may give rise to lymphatic and haematogenous metastases. The characteristics, depth of invasion, and extent of metastasis of the cancer largely influence a patient's life expectancy and are indicative of the effects and complications to be expected from the various therapeutic alternatives. The three factors are summarised in the *stage* of the cancer, which can be either I, IIA, IIB, III, IVA or IVB, in the order of progressive disease. To establish the stage of a patient's cancer, typically a number of diagnostic tests are performed, ranging from biopsies of the primary tumour to gastroscopic and endosonographic examinations of the oesophagus.

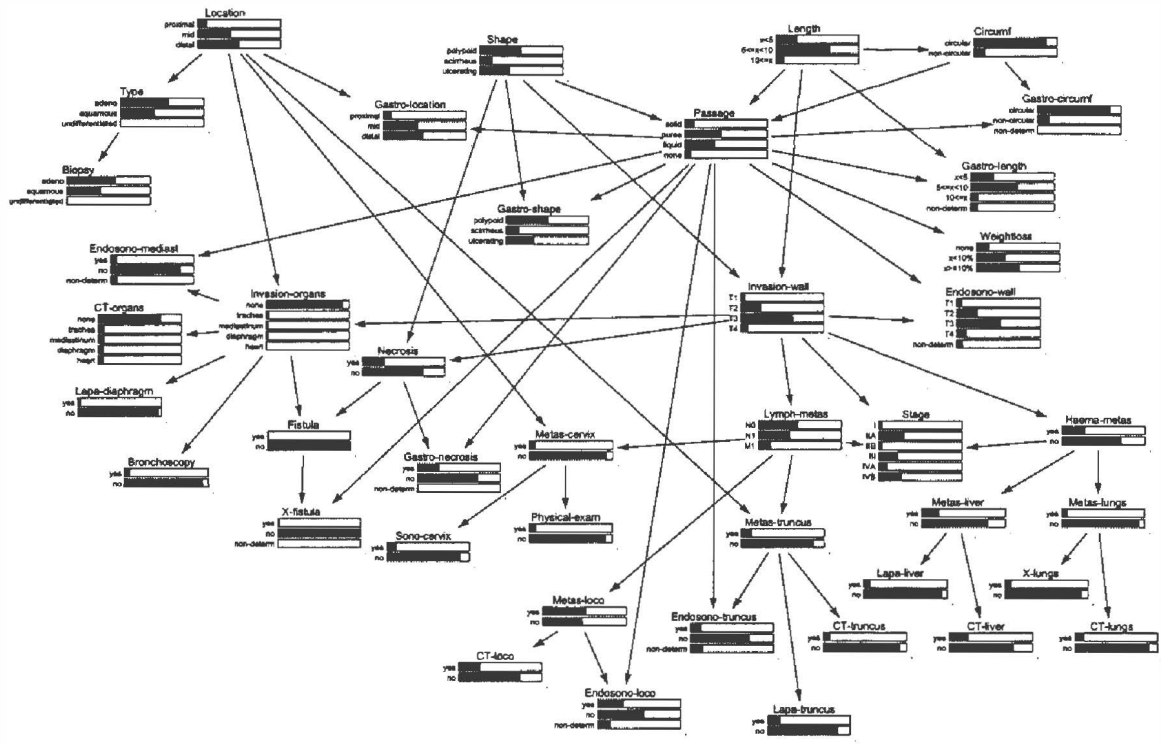

Figure 1: The oesophagus network.

The oesophagus network is depicted in Figure 1. It currently includes 42 variables. The number of values per variable ranges between two and six, with an average of 3.4. The number of incoming arcs per variable ranges between zero and three with an average of 1.7; the average number of outgoing arcs is 2.5, with a minimum of zero and a maximum of 11. For the network, a total of 932 probabilities are specified. The variable with the largest number of probabilities, 144, models the stage of the cancer; this variable is deterministic. The largest number of probabilities specified for a non-deterministic variable is 80.

Building upon the concepts outlined in Section 2, we performed a sensitivity analysis of the oesophagus network. We took the probability of a specific stage for the probability of interest. As six different stages are defined, we performed six separate analyses, each time focusing on another stage. Of the prior network, that is, the network without any evidence entered, we found 206 of the 932 parameters to be influential upon variation. For these parameters, the analyses yielded a total of 1236 sensitivity functions with two or three constants each. Because a network's sensitivities typically change with evidence, we repeated the analysis with data entered from 156 patients diagnosed with oesophageal cancer. The number of data entered per patient ranged between 6 and 21, with an average of 14.8. The parameters that we found to be influential upon variation differed between patients. The number of influential parameters also differed considerably and was found to be as high as 826.

## 4 The derivative and its sensitivity value

For extracting relevant information from the huge amount of sensitivity data that is typically generated from a probabilistic network, effective methods are needed. In this section, we discuss a method for this purpose that is based on the derivative of a sensitivity function and its associated sensitivity value. This method provides for selecting parameters that upon variation have a large effect on a probability of interest. In the next section, we introduce another method, based on the concept of admissible deviation, that focuses on the most likely value of a variable of interest.

In Section 2, we argued that the sensitivity of a probability of interest to variation of a parameter $x$ can be expressed as a function $f(x)$ of the form

$$f(x) = \frac{a \cdot x + b}{c \cdot x + d}$$

The first derivative $f'(x)$ of this function is

$$f'(x) = \frac{a \cdot d - b \cdot c}{(c \cdot x + d)^2}$$

The probability of interest is sensitive to deviations from the original assessment $x_0$ of the parameter under study, if the sensitivity value $|f'(x_0)|$ is greater than zero.

As an example, Figure 2 depicts a sensitivity function that we found for the oesophagus network. The function shows the effect of varying the parameter $p(\textit{Sono-cervix} = \textit{yes} \mid \textit{Metas-cervix} = \textit{no})$ on the probability $\Pr(\textit{Stage} = \text{III} \mid \textit{case } 6)$ for a specific patient. The assessment for the parameter under study is $x_0 = 0.07$. For this assessment, we find a sensitivity value of 0.53, which indicates that deviations from the original assessment would not greatly affect the probability of interest. If the assessment would have been 0.02 rather than 0.07, however, a much larger sensitivity value, of 4.74, would have been found, which indicates that even minor deviations would have had a considerable effect. For the value 0.20, on the other hand, a sensitivity value of just 0.07 is found. If the assessment for the parameter would have been 0.20, therefore, deviations would have had hardly any effect on the probability of interest.

Based upon the concept of sensitivity value, we might conclude that the parameters for whose assessments the sensitivity values are the largest, are the most likely to be quite influential upon variation and therefore should be selected for further investigation. Sensitivity values, however, provide insight in the effect of small deviations only from a parameter's assessment. When larger deviations are considered, the quality of the value as an approximation of the effect may break down rapidly, as noted before by Laskey [4]. For some parameters, it may be that the sensitivity values for slightly smaller assessments are very large and the sensitivity values for slightly larger assessments are quite small. For the parameter under study in Figure 2, for example, this property holds: the sensitivity value of the original assessment is rather small, but for smaller assessments especially the sensitivity values become quite large. Finding a relatively small sensitivity value for a parameter's assessment is therefore no guarantee that the probability of interest is hardly affected by variation of this parameter.

From our discussion of the sensitivity function shown in Figure 2, we have that the quality of a sensitivity value for indicating the effect of parameter variation decreases as the parameter's original assessment lies closer to the $x$-coordinate of the function's "*shoulder*". Therefore, to decide whether or not the probability of interest is likely to be affected by inaccuracies in the assessment for a specific parameter, we should consider not just the associated sensitivity value but also the distance of the assessment to the $x$-coordinate of the shoulder of the sensitivity function. We define the concept of shoulder more formally.

A sensitivity function is either a linear function or a hyperbola. For $c = 0$, for example, the sensitivity function is *linear*, as we then have that

$$f(x) = \frac{a \cdot x + b}{c \cdot x + d} = \frac{a \cdot x + b}{d} = \frac{a}{d} \cdot x + \frac{b}{d}$$

If the sensitivity function is not linear, it takes the form of a *hyperbola*:

$$f(x) = \frac{r}{x - s} + t,$$

with

$$r = \frac{b \cdot c - a \cdot d}{c^2}, \quad s = -\frac{d}{c}, \quad \text{and} \quad t = \frac{a}{c}$$

For ease of reference, Figure 3 depicts the general shape of a hyperbola. The hyperbola has two asymptotes, parallel to the $x$- and $y$-axes, in $x = s$ and $f(x) = t$, respectively. The hyperbola further is symmetrical in the line that has an absolute gradient of 1 and goes through the intersection point $(s, t)$ of the hyperbola's asymptotes. The point $(s, t)$ is termed the *center* of the hyperbola. The point at which the hyperbola intersects with the line in which it is symmetrical, is called the *vertex* of the hyperbola; this vertex is the point referred to before as the function's shoulder. As the symmetry line has a gradient with an absolute value of 1, the gradient of the hyperbola at the vertex also has an absolute value of 1. Using this property we can easily determine the $x$-coordinate $\hat{x}$ of the vertex:

$$|f'(x)| = \left| \frac{-r}{(x-s)^2} \right| = 1 \iff \hat{x} = s \pm \sqrt{|r|}$$

We now return to a sensitivity function $f(x)$ expressing a network's probability of interest in terms of a parameter $x$ under study. As noted above, if this function is not linear, it takes the form of a hyperbola. The center of the hyperbola is $(-\frac{d}{c}, \frac{a}{c})$. Because a sensitivity function is defined on the entire probability interval $[0,1]$ and moreover is nonnegative, we have that $-\frac{d}{c} < 0$ and $\frac{a}{c} \geq 0$. For the $x$-coordinate $\hat{x}$ of the vertex of the hyperbola, we find

$$\hat{x} = \frac{-d \pm \sqrt{|a \cdot d - b \cdot c|}}{c}$$

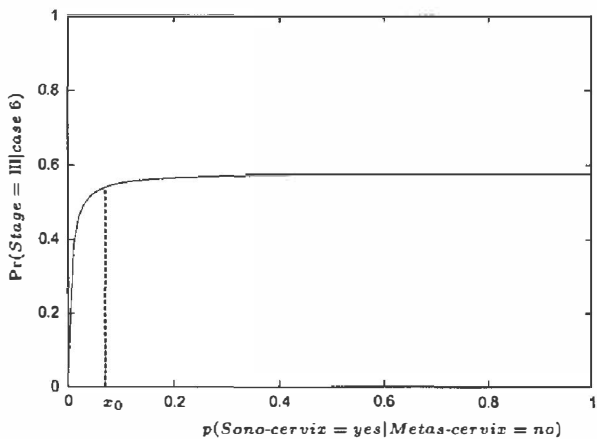

Figure 2: The sensitivity function $f(x)$ expressing the probability $\Pr(\textit{Stage} = \text{III} \mid \textit{case } 6)$ in terms of the parameter $x = p(\textit{Sono-cervix} = \textit{yes} \mid \textit{Metas-cervix} = \textit{no})$.

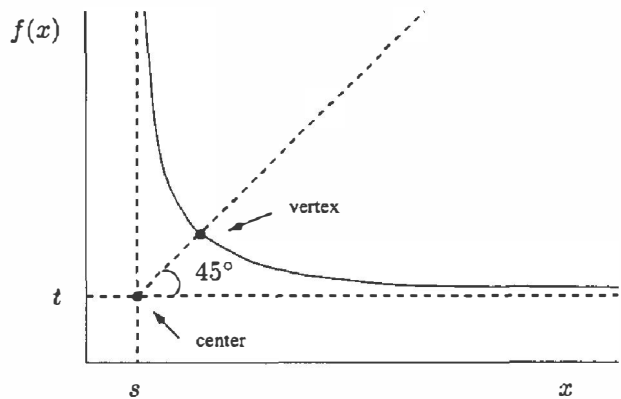

Figure 3: A hyperbola $f(x)$.

For the sensitivity function shown in Figure 2, for example, the $x$-coordinate of the vertex of the hyperbola is $\hat{x} = 0.05$. We observe that the original assessment $x_0 = 0.07$ for the parameter under study lies quite close to this coordinate. We recall that the sensitivity value for $x_0$ is just 0.53. Because the assessment lies close to $\hat{x}$, however, the sensitivity value cannot be considered a good approximation of the effect of the parameter's variation: small deviations from the assessment, especially to smaller values, have a considerable effect on the probability of interest. We conclude that this parameter should be selected for further investigation, regardless of its small sensitivity value.

From the above observations, we have that for extracting from the sensitivity data the parameters that deserve further attention, parameters whose assessment has a sensitivity value larger than 1 should be identified. In addition, parameters whose original assessment lies close to the $x$-coordinate of the vertex of the associated sensitivity function should be selected.

To conclude our discussion of sensitivity values, we would like to note that in the analysis of the oesophagus network we found rather strong sensitivities. An example is shown in Figure 4: for the parameter under study, a sensitivity value of 6.97 was found.

## 5 Admissible deviation

In the previous section, we focused on the derivative of a sensitivity function and showed how its associated sensi-tivity value can be used for selecting parameters that upon variation have a large effect on a specific probability of interest. Often, however, we are interested not in the effect of parameter variation on a probability of interest, but in the effect on the most likely value for a variable of interest. In a medical application, for example, the most likely diagnosis given a patient's symptoms and signs may be the outcome of interest. For this type of outcome, the derivative of a sensitivity function and its associated sensitivity value are no longer appropriate for establishing a parameter's effect upon variation. For some parameters, deviation from their original assessment may have a considerable effect on the probability of a specific outcome and yet not induce a change in the most likely one; for other parameters, variation may have little effect on the probabilities involved and nonetheless result in a change in the most likely outcome. We would like to note that the idea of focusing on the most likely value of a specific variable conforms to the practice of sensitivity analysis of decision-theoretic models where the most preferred decision is the outcome of interest [9].

To provide for studying the effects of parameter variation in view of a most likely outcome, we enhance the basic method of sensitivity analysis for probabilistic networks with the computation of an interval within which a parameter can be varied without inducing a change in the most likely value of a variable of interest. Now, let $A$ be the variable of interest. Let $x$ be the parameter under study and let $x_0$ be its assessment. The *admissible deviation* for $x_0$ is a pair of real numbers $(r, s)$ such that $x_0$ can be varied between the bounds $\max(x_0 - r, 0)$ and $\min(x_0 + s, 1)$ without inducing a change in the most likely value for the variable $A$; $r$ and $s$, moreover, are the largest numbers for which this property holds. Note that the interval within which a parameter can be freely varied can be bounded for two reasons: either the most likely outcome changes if the parameter is varied beyond the specified boundary, or the boundary of the probability interval [0, 1] has been reached. To express that a parameter can be varied as far as the boundary of the probability interval, we use the symbol $\infty$ in the admissible deviation associated with its assessment.

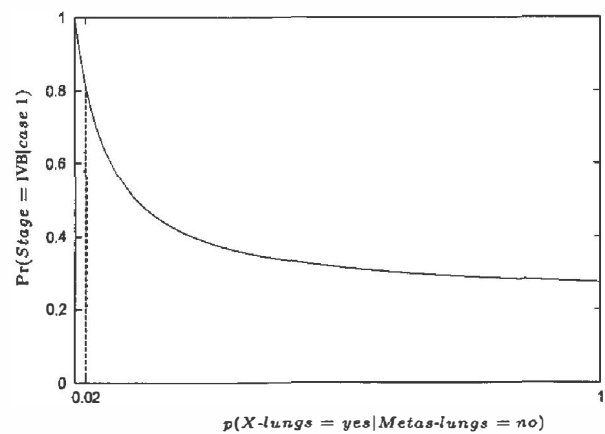

Figure 4: The sensitivity function $f(x)$ expressing the probability $\Pr(\textit{Stage} = \text{IVB} \mid \textit{case } 1)$ in terms of the parameter $x = p(\textit{X-lungs} = \textit{yes} \mid \textit{Metas-lungs} = \textit{no})$.

The admissible deviation for a parameter's assessment can be computed from the sensitivity functions $f_{\Pr(a_i|e)}$ that are yielded for the various values $a_i$ of the variable $A$. More specifically, the admissible deviation is established by studying the points at which two or more of these functions intersect. Suppose that for the original assessment $x_0$ of the parameter $x$, the value $a_i$ is the most likely value for the variable of interest. Also suppose, for ease of exposition, that neither of the bounds of the admissible deviation equals $\infty$. The admissible deviation for $x_0$ now is the pair $(r, s)$, where the leftmost deviation $r$ is the smallest real number for which $x_0 - r$ is the $x$-coordinate of a point at which the sensitivity function $f_{\Pr(a_i|e)}(x)$ intersects with another sensitivity function $f_{\Pr(a_j|e)}(x)$ with $j \neq i$; the rightmost deviation $s$ is defined analogously.

We present various examples to illustrate the difference between using sensitivity values and admissible deviations for studying the effects of parameter variation. The examples are taken from patient-specific analyses of the oesophagus network. The figures used display the sensitivity functions yielded for all the values of the variable *Stage* that models the stage of a patient's cancer. As we are now considering a variable of interest rather than a probability of interest, using the concept of sensitivity value for selecting interesting parameters amounts to examining the sensitivity values from *all* functions for the parameter under study.

Our first example addresses a parameter that induces small sensitivity values and upon variation does not result in a change in the most likely outcome. The sensitivity functions for the parameter are depicted in Figure 5: the figure shows the effects of varying $x = p(CT\text{-}lungs = yes \mid Metas\text{-}lungs = no)$ on the probabilities of the various different values of the variable *Stage* for a specific patient. From the figure, it is readily seen that the parameter's assessment has associated a small sensitivity value from each of the sensitivity functions. The sensitivity function $f_{\text{IVB}}(x)$, for example, that expresses the probability $\Pr(Stage = \text{IVB} \mid case\ 138)$ in terms of the parameter under study, is

$$f_{\text{IVB}}(x) = \frac{0.09208 \cdot x + 1.17403}{x + 1.17403}$$

from which we find a sensitivity value of

$$|f'_{\text{IVB}}(0.05)| = 0.71145$$

We further observe that the assessment $x_0 = 0.05$ for the parameter lies relatively far from the $x$-coordinate $\hat{x}$ of the function's vertex:

$$\begin{aligned}\hat{x} &= \frac{-1.17403+\sqrt{|0.09208 \cdot 1.17403 - 1.17403 \cdot 1|}}{1} \\ &= -0.14\end{aligned}$$

Based on the concept of sensitivity value as discussed in the previous section, we conclude that the parameter does not deserve additional attention. Inspection of Figure 5 now further reveals that the admissible deviation for the parameter's assessment equals $(\infty, \infty)$, which indicates that the parameter can be varied over the entire probability interval $[0, 1]$ without inducing a change in the most likely stage. Based on the concept of admissible deviation, therefore, the parameter should also be further disregarded.

For our next example, we consider a parameter that induces large sensitivity values, but upon variation is not expected to result in a change in the most likely outcome. The sensitivity functions for the parameter are depicted in Figure 6: the figure shows the effects of varying $x = p(CT\text{-}loco = yes \mid Metas\text{-}loco = no)$ on the probabilities of the dif-ferent values of the variable *Stage*. For the parameter's original assessment $x_0 = 0.02$, large sensitivity values are found from some of the six functions. For example, from the sensitivity function that expresses the probability $\Pr(Stage = \text{IIA} \mid case\ 82)$ in terms of the parameter, a sensitivity value of 2.07 is found. When studying the effects of variation on the probabilities of interest, therefore, the parameter should be selected for further investigation. Now consider the effects of variation on the most likely value of the variable *Stage*. For the parameter's assessment $x_0$, we find that stage III is the most likely stage for the patient under study. The sensitivity function for this stage intersects with the sensitivity function for stage IIA at $x = 0.17$. Further inspection of the figure reveals that the admissible deviation for the parameter's assessment is $(\infty, 0.15)$. We observe that the rightmost deviation is relatively large compared to the original assessment. We therefore conclude that inaccuracies in the assessment are not reasonably expected to affect the most likely outcome, and the parameter should not be selected for further investigation.

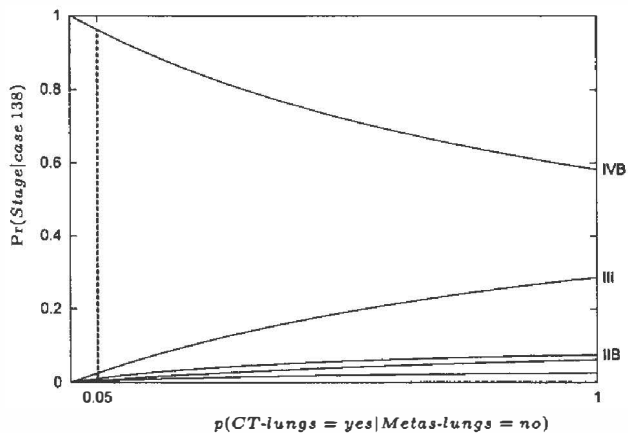

Figure 5: The sensitivity functions expressing the probabilities Pr(*Stage* | *case* 138) in terms of the parameter $p(CT\text{-}lungs = yes \mid Metas\text{-}lungs = no)$.

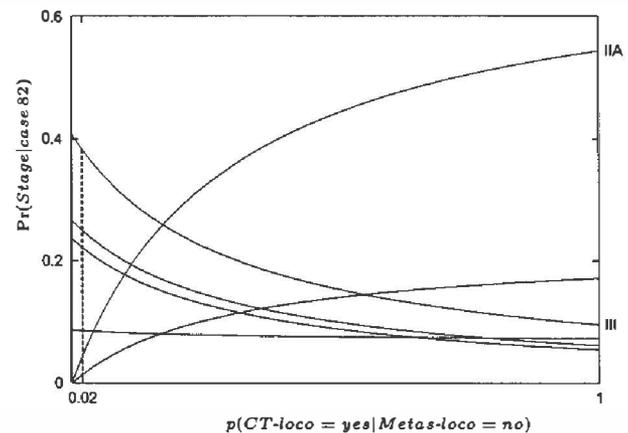

Figure 6: The sensitivity functions expressing the probabilities Pr(*Stage* | *case* 82) in terms of the parameter $p(CT\text{-}loco = yes \mid Metas\text{-}loco = no)$.

We now address a parameter that induces small sensitivity values but upon variation affects the most likely outcome. Figure 7 depicts the sensitivity functions for the parameter $x = p(Wall\text{-}inv = T2 \mid Shape = polypoid, 5 \leq Length < 10cm)$. For the original assessment $x_0 = 0.2$ of the parameter, the largest sensitivity values are found from the functions for the stages IIA and III:

$$|f'_{\text{IIA}}(0.2)| = 0.32151$$
$$|f'_{\text{III}}(0.2)| = 0.34832$$

As the sensitivity functions are linear in the parameter under study, the computed sensitivity values describe the effects of variation exactly. Based upon the concept of sensitivity value, we therefore conclude that the parameter should not be selected for further investigation. Now, the admissible deviation for the parameter's assessment is computed to be $(0.10, \infty)$. As the leftmost deviation is quite small compared to the assessment 0.20, we find that the parameter deserves further investigation. We would like to note that, if the original assessment had been in the interval $[0.04, 0.10]$, then the sensitivity values would not have changed, but the admissible deviation would have become extremely small for both directions of variation.

Our final example pertains to a parameter that induces large sensitivity values and upon variation is expected to result in a change in the most likely outcome. The sensitivity functions for the parameter are displayed in Figure 8: the figure shows the effects of varying $x = p(CT\text{-}organs = mediast \mid Inv\text{-}organs = none)$. From two of the sensitivity functions, we find relatively large sensitivity values:

$$|f'_{\text{IIA}}(0.05)| = 1.34333$$
$$|f'_{\text{III}}(0.05)| = 0.99446$$

In addition, the original assessment $x_0 = 0.05$ for the parameter lies very close indeed to the $x$-coordinates $\hat{x}_{\text{IIA}}$ and $\hat{x}_{\text{III}}$ of the vertices of the two functions:

$$\hat{x}_{\text{IIA}} = 0.0597$$
$$\hat{x}_{\text{III}} = 0.0499$$

Based upon the concept of sensitivity value, therefore, the parameter should be selected for further investigation. We now consider the effects of variation on the most likely outcome. We observe that for the original assessment the most likely stage is IIA. The admissible deviation for the assessment is computed to be $(0.002, \infty)$. We observe that the leftmost deviation is extremely small, indicating that an inaccuracy by just 0.002 in the original assessment will lead to a different most likely value for the variable *Stage* and may in fact result in a different treatment decision for the patient under consideration.

The above examples illustrate that admissible deviations provide additional insight into the sensitivity of a probabilistic network's outcome to inaccuracies in its parameters. We would like to note that the use of admissible deviations is especially of interest when decisions are to be based on the most likely value of a variable of interest [10].

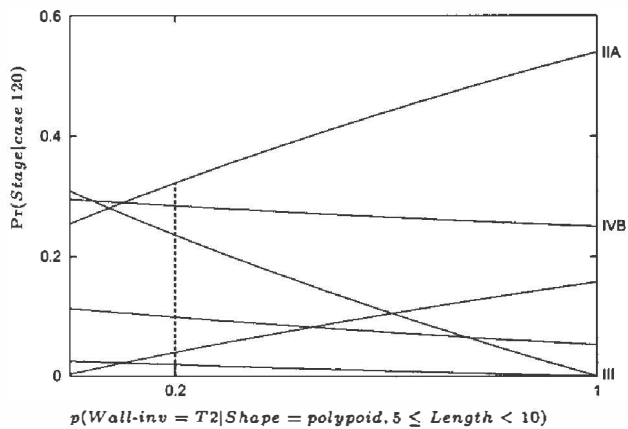

Figure 7: The sensitivity functions expressing the probabilities $\Pr(Stage \mid case\ 120)$ in terms of the parameter $p(Wall\text{-}inv = T2 \mid Shape = polypoid, 5 \leq Length < 10cm)$.

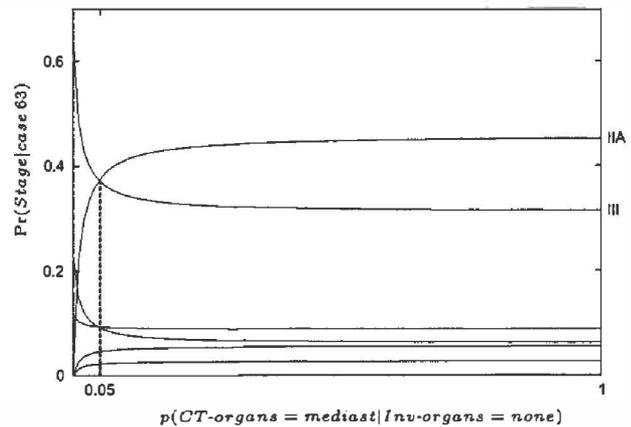

Figure 8: The sensitivity functions expressing the probabilities $\Pr(Stage \mid case\ 63)$ in terms of the parameter $p(CT\text{-}organs = mediast \mid Inv\text{-}organs = none)$.

## 6 Conclusions

Sensitivity analysis of a probabilistic network serves to yield insight in the effects of inaccuracies in its numerical parameters. As for real-life networks a sensitivity analysis tends to result in a huge amount of data, effective methods for selecting relevant parameters from the generated data are called for. In this paper, we introduced two such methods.

One of our methods for extracting relevant information from sensitivity data builds upon the derivative of a sensitivity function and its associated sensitivity value. Basically, a parameter $x$ is selected for further investigation if the sensitivity function that expresses the network's probability of interest in terms of $x$, has a large sensitivity value, that is, a large absolute gradient at the original assessment $x_0$ for the parameter. We argued that the sensitivity value of a parameter's assessment does not constitute a sufficient criterion for its selection: if the assessment lies close to the $x$-coordinate of the vertex of the associated sensitivity function, then small deviations may have a large effect on the probability of interest, regardless of the assessment's sensitivity value. To identify parameters that are likely to have a large effect on the probability of interest, therefore, not only the sensitivity value of its assessment need to be considered, but also the distance of the assessment to the $x$-coordinate of the sensitivity function's vertex.

The derivative and its associated sensitivity value are suitable for identifying parameters that have the largest influence on a network's probability of interest. We argued, however, that if the network's outcome is not a single probability, but rather the most likely value for a variable of interest, then the sensitivity value is no longer appropriate for this purpose. We introduced the concept of admissible deviation to provide for studying the effects of parameter variation on this type of outcome: the admissible deviation $(r, s)$ for a parameter's assessment indicates the largest interval within which the assessment can be varied freely without inducing a change in the most likely outcome.

We illustrated the various concepts introduced in this paper by means of a sensitivity analysis of the oesophagus network, a moderately-sized real-life network in the field of oncology. The analysis of this network resulted in a huge amount of data in which considerable sensitivities were hidden. Our methods proved quite useful for extracting the relevant information from these data.

**Acknowledgements**. This research has been (partly) supported by the Netherlands Computer Science Research Foundation with financial support from the Netherlands Organisation for Scientific Research (NWO). We are most grateful to Babs Taal and Berthe Aleman from the Netherlands Cancer Institute, Antoni van Leeuwenhoekhuis, who spent much time and effort in the construction of the oesophagus network.